\documentclass[10pt,conference]{IEEEtran}
\IEEEoverridecommandlockouts

\usepackage{cite}
\usepackage{subfig}
\usepackage{amsmath,amssymb,amsfonts}
\usepackage{algorithmic}
\usepackage{graphicx}
\usepackage{enumitem}
\usepackage{textcomp}
\usepackage{xcolor}
\usepackage{tikz, pgfplots, csvsimple}
\usepackage{hyperref}

\def\BibTeX{{\rm B\kern-.05em{\sc i\kern-.025em b}\kern-.08em
    T\kern-.1667em\lower.7ex\hbox{E}\kern-.125emX}}

\begin{document}
\title{Improving Explainability of Image Classification in Scenarios with Class Overlap: Application to COVID-19 and Pneumonia}

%\iffalse
\author{
  \IEEEauthorblockN{%
    Edward Verenich\IEEEauthorrefmark{3}\,\IEEEauthorrefmark{1},
    Alvaro Velasquez\IEEEauthorrefmark{1},
    Nazar Khan\IEEEauthorrefmark{2},
    Faraz Hussain\IEEEauthorrefmark{3}
  }\vspace{1.5ex}
  \IEEEauthorblockA{%
    \IEEEauthorrefmark{3}Clarkson University, Potsdam, NY\\{\it \{verenie, fhussain\}@clarkson.edu}\vspace{1.5ex}\\
    \IEEEauthorrefmark{2}Punjab University College of Information Technology\\{\it nazarkhan@pucit.edu.pk}\vspace{1.5ex}\\
    \IEEEauthorrefmark{1}Air Force Research Laboratory, Rome, NY\\{\it \{edward.verenich.2, alvaro.velasquez.1\}@us.af.mil}
  }
}
%\fi

\maketitle

\begin{abstract}
Trust in predictions made by machine learning models is increased if the model generalizes well
on previously unseen samples and when inference is accompanied by cogent explanations of the reasoning
behind predictions.
In the image classification domain, generalization can be assessed through accuracy, sensitivity,
and specificity. 
Explainability can be assessed by how well the model localizes the object of interest within an image.
However,  both generalization and explainability through localization are degraded
in scenarios with significant overlap between classes.
We propose a method based on binary expert networks that enhances the explainability
of image classifications through better localization by mitigating the model uncertainty induced by class overlap.
Our technique performs discriminative localization on images that contain features with significant class overlap,
without explicitly training for localization.
Our method is particularly promising in real-world class overlap scenarios, such as COVID-19 and pneumonia,
where expertly labeled data for localization is not readily available.
This can be useful for early, rapid, and trustworthy screening for COVID-19.
\end{abstract}

\begin{IEEEkeywords}
  explainability, trust in AI, class overlap, COVID-19, data-starved, deep learning,
  object localization.
\end{IEEEkeywords}

\section{Introduction}
The use of deep neural networks for image classification and object detection in imagery is well established in the computer vision domain.
As neural networks became increasingly used in real world applications,
such as assisting medical diagnosis, the phenomenon of class overlap became more apparent \cite{alejo2013hybrid}.
Recent work on detecting COVID-19 %(previously known as coronavirus-infected pneumonia (NCIP))
using X-ray imagery has also shown that class overlap degrades classifier performance  \cite{OZTURK2020103792}.
By training  convolutional neural networks (CNNs)
to account for classes with similar conditions, the model becomes less certain.
This is in part due to overlap in class activations triggered by the same image.
This paper presents {\em a new technique to distinguish between COVID-19 and regular pneumonia in X-ray imagery
  in a more explainable fashion by using class activation maps}.

\begin{figure}[t]
 \centering
 \fbox{\includegraphics[width=\linewidth]{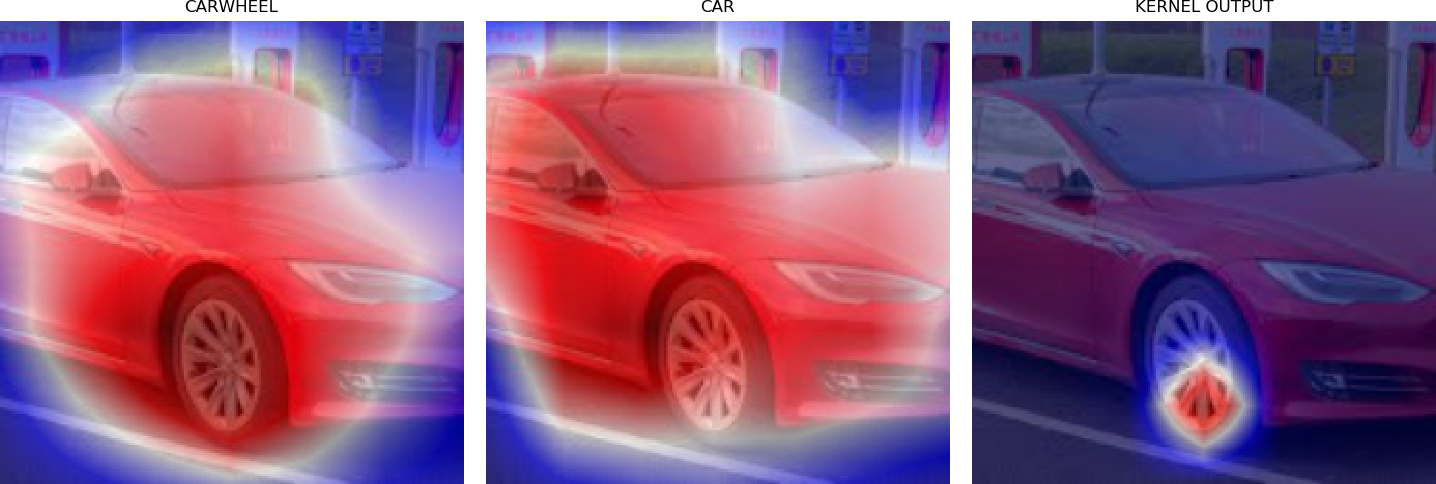}}
    \caption{The left and center images show significant overlap in the class activation maps computed by
      the binary classifiers for carwheels and cars, respectively.
      The third image, computed by applying our kernel function
      on the  first two images,
      localizes the region  in the original image  that is 
      primarily responsible for its classification  as a carwheel.
    }
    \label{fig:example}
\end{figure}

The standard approach in deep learning to reduce uncertainty is to provide more training data to the model, which is not always possible and primarily addresses model uncertainty that is due to model parameters. Another method to reduce decision uncertainty is to localize target objects, hence increasing confidence in the prediction.
Localization through supervised training with labeled bounding boxes
to compute the reward is a widely used 
approach for reducing decision uncertainty in image classification~\cite{long2017accurate}.
However, the lack of labeled data and inherent noise present in novel situations results in additional
predictive uncertainty.
Consider, for example, X-ray imagery of confirmed COVID-19 patients,
where X-ray images were taken to analyze pulmonary complications,
yet expert localization of COVID-19 specific attributes was not
performed by radiologists,
i.e. no bounding boxes on X-ray images of COVID-19 relevant regions were annotated
\cite{wang2020covid, cohen2020covid, chungFig1CovidXray, chungActualmedCovid, Chowdhury_2020}.

\begin{figure*}[t!]
  \centering
  \fbox{\includegraphics[width=0.85\linewidth]{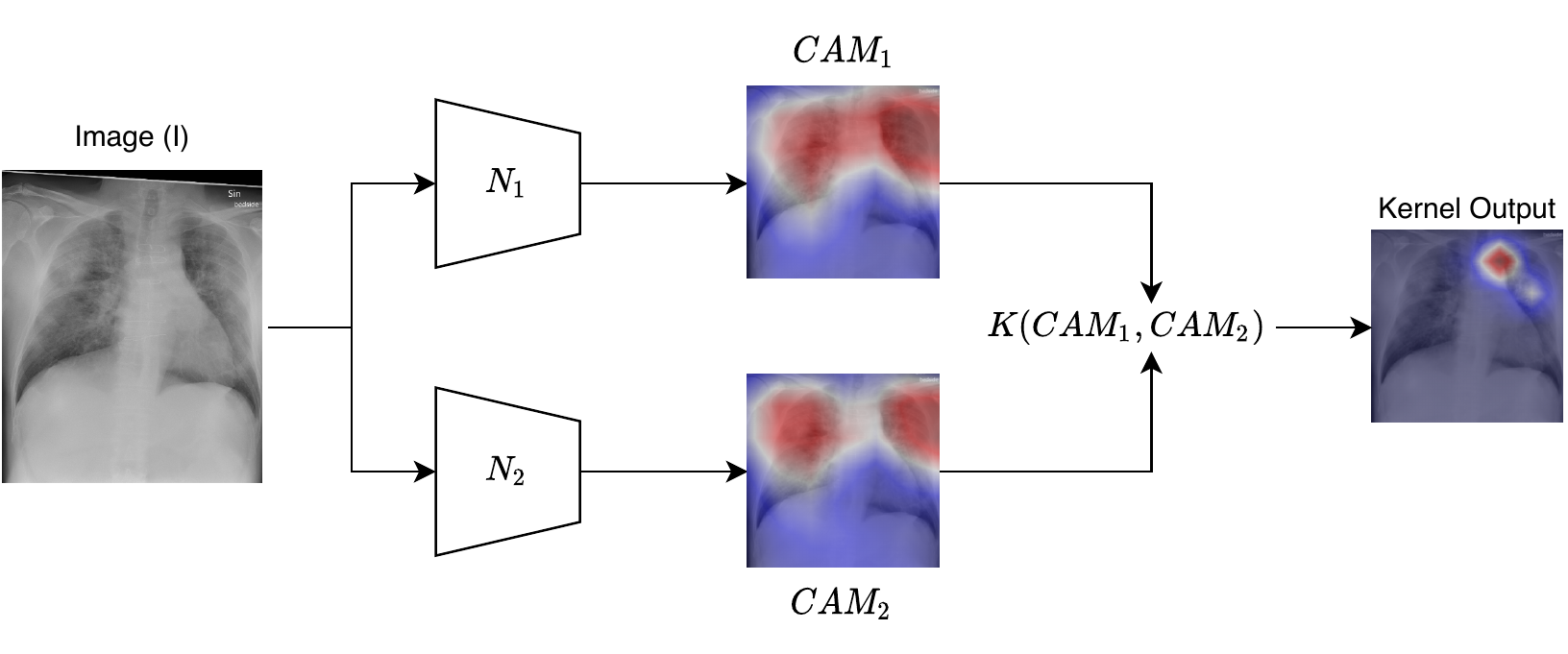}}
  \caption{High-level view of our dual-network technique for handling class overlap.
    Given a class of interest $C_1$ and another possibly overlapping class $C_2$,
    our approach involves training two separate binary expert networks ($N_1$, $N_2$).
    Each input image ($I$) is fed to both expert networks to obtain class activation maps ($CAM_1$, $CAM_2$)
    which are then used by our directed kernel function $K$ to localize regions in $I$ where 
   the expert network for the class of interest ($N_1$) is more confident.}
  \label{fig:archdiag}
\end{figure*}

Ghoshal et al.~\cite{Ghoshal2020EstimatingUA} note that there are two distinct kinds of predictive uncertainty in deep learning.
First, \textit{epistemic} uncertainty, or uncertainty in model parameters, which decreases
with more training data.
Second, \textit{aleatoric} uncertainty, that accounts for noise in observations due to class overlap, label noise, and varying error term size across values of an independent variable.
Aleatoric uncertainty cannot be easily reduced by increasing the size of the training set.

At the heart of our approach is the use of class activation maps (CAMs) for improved localization
of the regions  responsible for the image being in a specific class (e.g. COVID-19)
as opposed to some other overlapping class (pneumonia).
As an example, \autoref{fig:example} shows activation maps for two separate models trained
for classifying \textit{carwheels} and \textit{cars}, respectively.
The figure depicts significant overlap in the regions responsible for 
categorizing the image in the two classes.
{\em Our goal is localizing regions in the image  which are more responsible for its classification
in a specific class of interest (carwheel), in particular where bounding boxes are not available during training.}

In unexpected public health emergencies, such as the coronavirus pandemic, 
labeled datasets with bounding box annotations are unlikely to be available to
the community at an early stage.
Such scenarios preclude the possibility of training a model for localization.
In such situations, is it possible to improve localization when there is no way to train for it? 

This work proposes a method for improved localization  in order to enhance
explainability of image classifications in data regimes with significant class overlap,
thus mitigating aleatoric uncertainty. 
Our results show that 
{\em training per-class binary CNN models  and
 applying our new kernel function on their class activation maps 
  can extract and better localize objects from overlapping classes}.

\section{Related Work}
Image classification and object localization have been successfully utilized for diagnostic purposes in radiology for pneumonia detection using chest X-rays~\cite{ChexNet:DBLP:journals/corr/abs-1711-05225}. With the recent emergence of the COVID-19,
a number of methods and models have been proposed, as surveyed by Shi et al.~\cite{Shi:article},
to detect the disease using medical imaging.
Wang and Wong~\cite{wang2020covid} released early work on using convolutional neural networks for COVID-19 detection from X-ray images.  Alqudah et al.~\cite{Alqudah2020Detection} used convolutional neural networks to classify X-ray images as well as extract features and pass them to other classifiers. Ghoshal et al.~\cite{Ghoshal2020EstimatingUA} observed that most methods focused exclusively on increasing accuracy without accounting for uncertainty in the decision and proposed a method to estimate decision uncertainty. Zhou et al.~\cite{zhou2016learning} showed that discriminative localization is possible without explicitly training for object detection with labeled object bounding boxes. However, noise related to overlapping classes has not been considered in these works. This work builds on these ideas and extends the work of Zhou et al.~\cite{zhou2016learning} on discriminative localization to specifically address the problem of overlapping classes and explainability in image classification.

\section{Approach}
\begin{figure*}[t]
    \centering
    \includegraphics[width=0.9\linewidth]{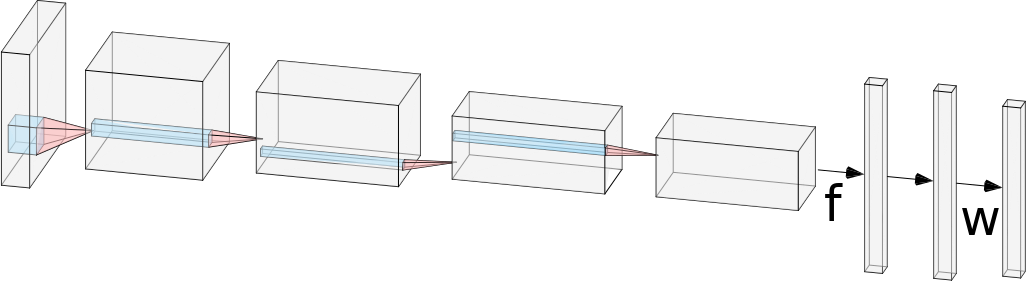}
    \caption{Example ResNet based expert model trained as a binary classifier.
      The second to fourth boxes from the left represent the four convolutional stages of the ResNet architecture shared by various ResNet implementations. Activations $f$ are extracted before the Global Average Pooling layer, and class weights $w$ come from the connections between the flatten layer and the final Fully Connected layer. The dimensions of the last three layers starting from the left (following $f$) are 1x1x2048, 2048, and 2 for binary classification. }
    \label{fig:resnet}
\end{figure*}

Our goal is to  localize image regions responsible for a class of interest (e.g. COVID-19),
given possibly overlapping classes (e.g. pneumonia/COVID-19).
Our method consists of training locally independent expert networks as binary classifiers
for two different classes that are possibly overlapping,
e.g. classifiers for COVID-19/No-COVID-19 and pneumonia/No-pneumonia. %non-SARS-CoV2
These binary expert networks are then leveraged as expert classifiers on each input image
as part of a dual-network architecture as shown in \autoref{fig:archdiag}.
CAMs obtained from the two networks are then passed to our novel kernel function ($K$)
that localizes image regions responsible for the class of interest.
Note that, in our approach, a binary expert network is a classifier that classifies
its input as either in-class or out-of-class for a specific class
e.g. a COVID-19/No-COVID-19 classifier is a binary expert network for COVID-19.
Similarly, a Pneumonia/No-Pneumonia classifier is an expert network for Pneumonia.
However, we do not consider a COVID-19/Pneumonia classifier as a binary expert network.
\newline

 \noindent Our approach is summarized below:
\begin{enumerate}[topsep=3pt,itemsep=0ex,partopsep=1ex,parsep=1ex,label={\arabic*.}] 
\item Given a class of interest $C_1$  and a possibly overlapping class $C_2$,
  train separate binary expert networks ($N_1, N_2$) for both of them.
    \item Pass each input image through $N_1$ and $N_2$ and, in both cases,
      the features of the last convolutional layer are extracted as CAMs ($CAM_1, CAM_2$).
    \item $CAM_1$ and $CAM_2$ are passed to our kernel function which localizes regions in the image
      classifier ($N_1$) of the class of interest is more confident.
\end{enumerate}

\noindent We now describe how the CAMs are computed and then define our kernel function.

\subsection{Class Activation Maps}
Class activation maps allow us to localize objects of a given class by mapping regions of an image to the most active values in the activation layer of a network. To obtain CAMs of each of the binary expert classifiers we follow the approach described by Zhou et al.~\cite{zhou2016learning}, differing only in the architecture of the expert models. In order to compute a CAM, a convolutional network architecture needs an activation layer, followed by a pooling layer (average or max), and a fully connected layer to obtain a class score.
A ResNet architecture meets these requirements, utilizing a Global Average Pooling layer
after the final convolutional layer.  

We create a mechanism for extracting values from the Activation layer, in case of our ResNet based networks it is the last layer before the Global Average Pooling layer. We add a hook to the Activation layer to store its activation values when a forward pass is performed on a network for inference. The activation values in the Activation layer contain 2048 activation maps $[f_1 ... f_{2048}]$ , each with a dimension of 7x7, thus when we extract activation values after a forward pass we get a tensor with a shape of 7x7x2048. Within the normal forward pass, this tensor is then reduced by the Global Average Pooling layer to a tensor of shape 1x1x2048 by averaging each feature map and then flattening the tensor to just 2048 in the Flatten layer. The connections between the Flatten layer and final Fully Connected Layer contain the weights $[w_1 ... w_i]$ that are used for classification, as each node in the final connected layer represents an object class. We use these weights $w_i$ along with activation maps $f_i$ to compute a CAM for a predicted class during a forward pass. \autoref{fig:resnet} shows a high level view of an expert network and locations of activation ($f$) and weight ($w$) tensors, respectively. Finally we compute a weighted sum using activation values and weights for a predicted class as in \autoref{eq:cam_sum} to produce a 7x7 tensor
that represents a CAM for the predicted class $c$.

\begin{equation}\label{eq:cam_sum}
   CAM_{c} = w^c_1 \cdot f_1 + w^c_2 \cdot f_2 + ... + w^c_{2048} \cdot f_{2048}
\end{equation}

In order to identify regions in the image that are important to the predicted class, we superimpose the 7x7 CAM on the original image using bilinear sampling to scale the CAM to appropriate size, which in this case is 224x224.

\begin{figure*}[!th]
    \centering
    \subfloat[]{%
      \fbox{\includegraphics[width=0.90\columnwidth]{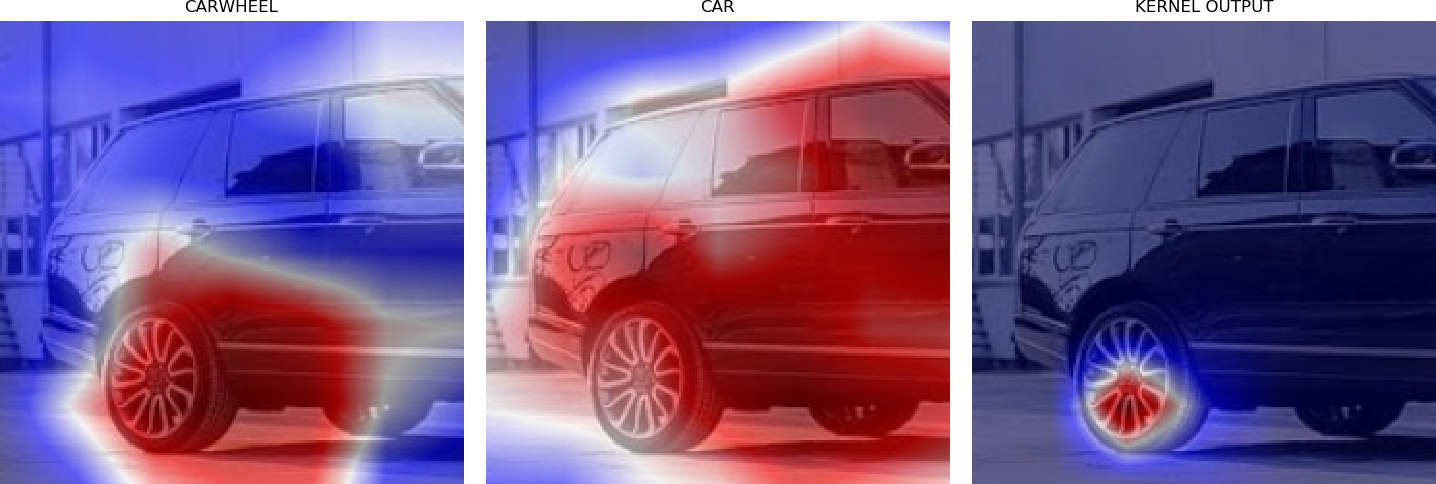}}%
    }
    \hspace{0.1in}
     \subfloat[]{%
       \fbox{\includegraphics[width=0.90\columnwidth]{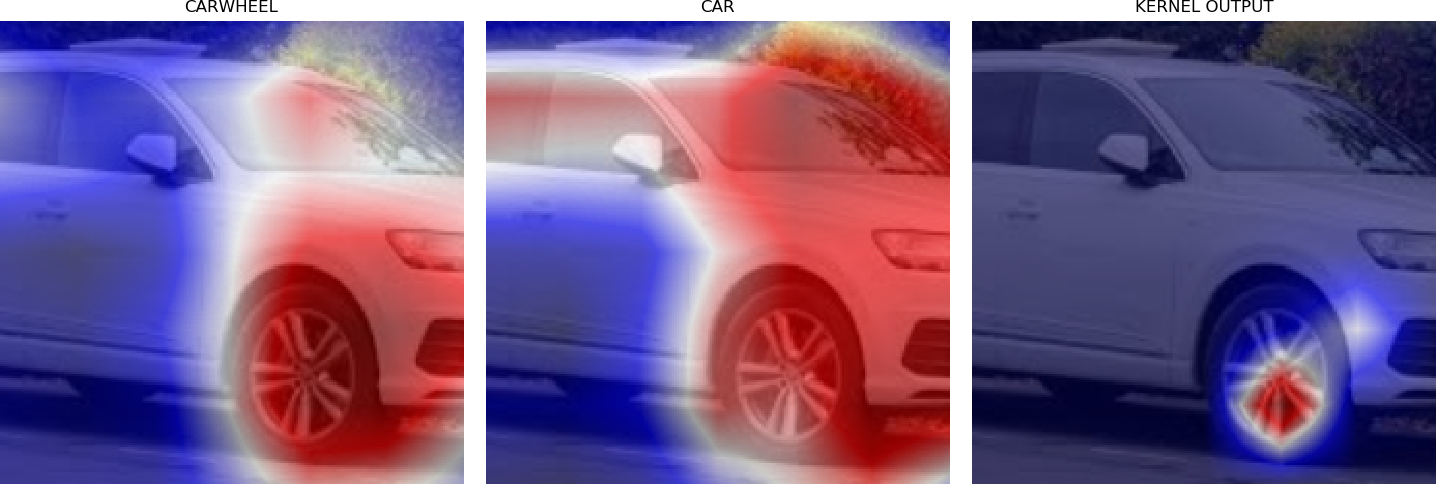}}%
    }\\
    \vspace{0.2in}

     \subfloat[]{%
       \fbox{\includegraphics[width=0.90\columnwidth]{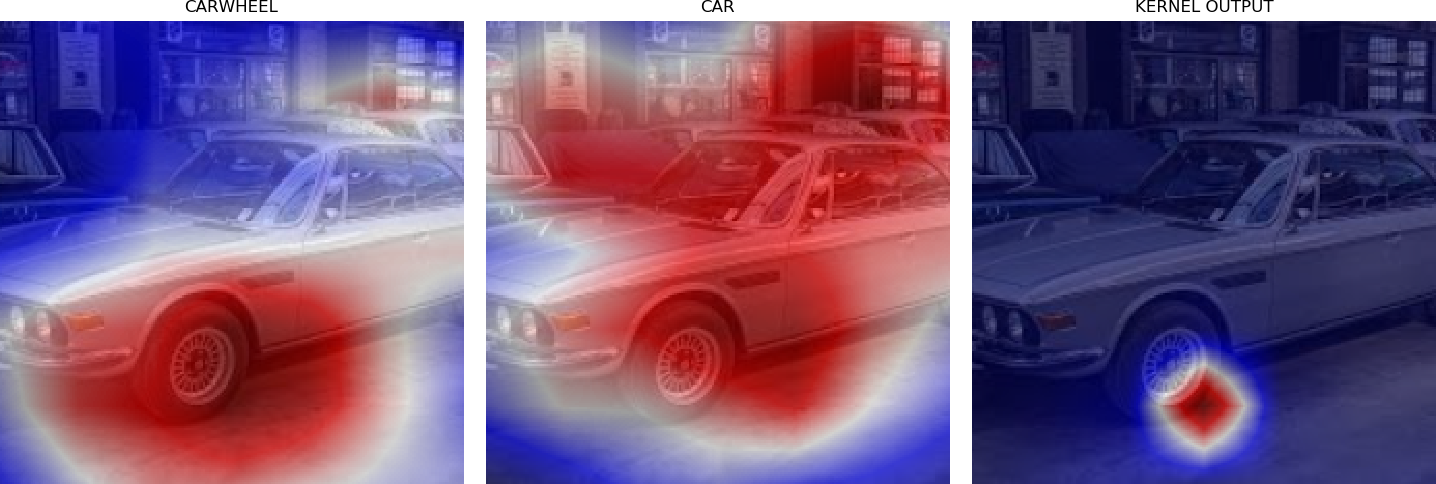}}%
     }
     \hspace{0.1in}
     \subfloat[]{%
       \fbox{\includegraphics[width=0.90\columnwidth]{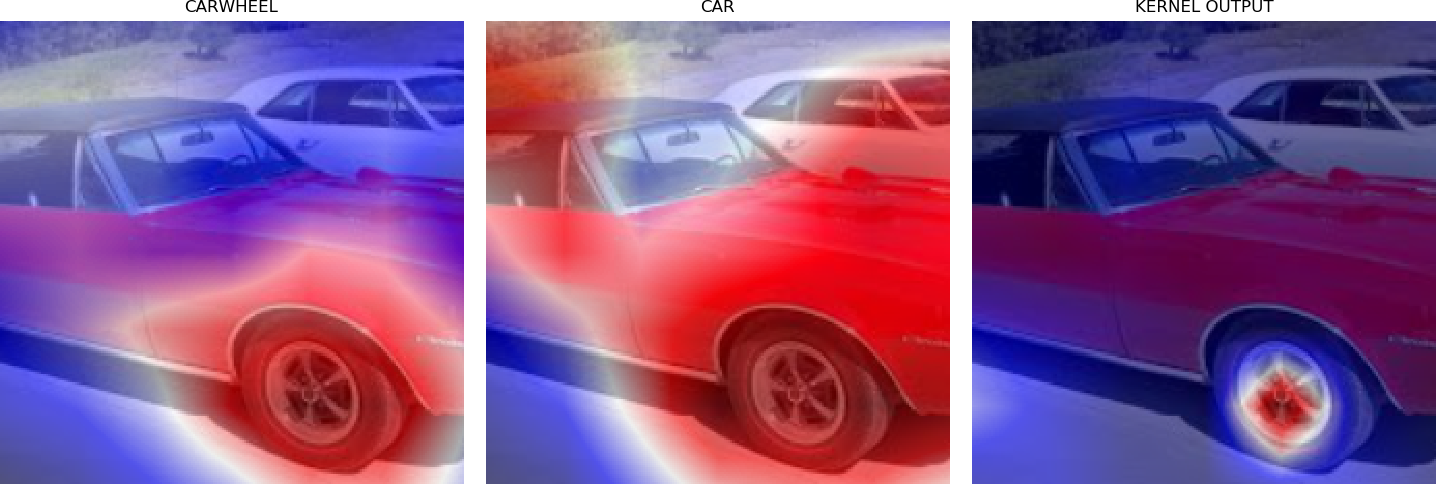}
       }
       }

       \vspace{0.2in}
     \subfloat[]{%
       \fbox{\includegraphics[width=0.90\columnwidth]{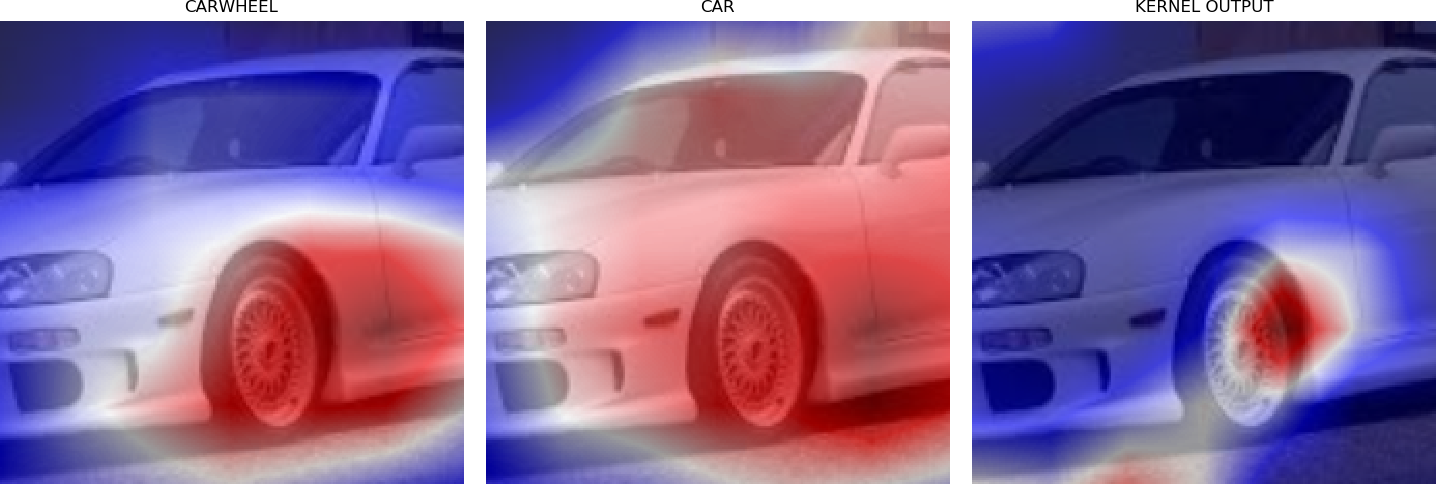}}%
     }
     \hspace{0.1in}
     \subfloat[]{%
       \fbox{\includegraphics[width=0.90\columnwidth]{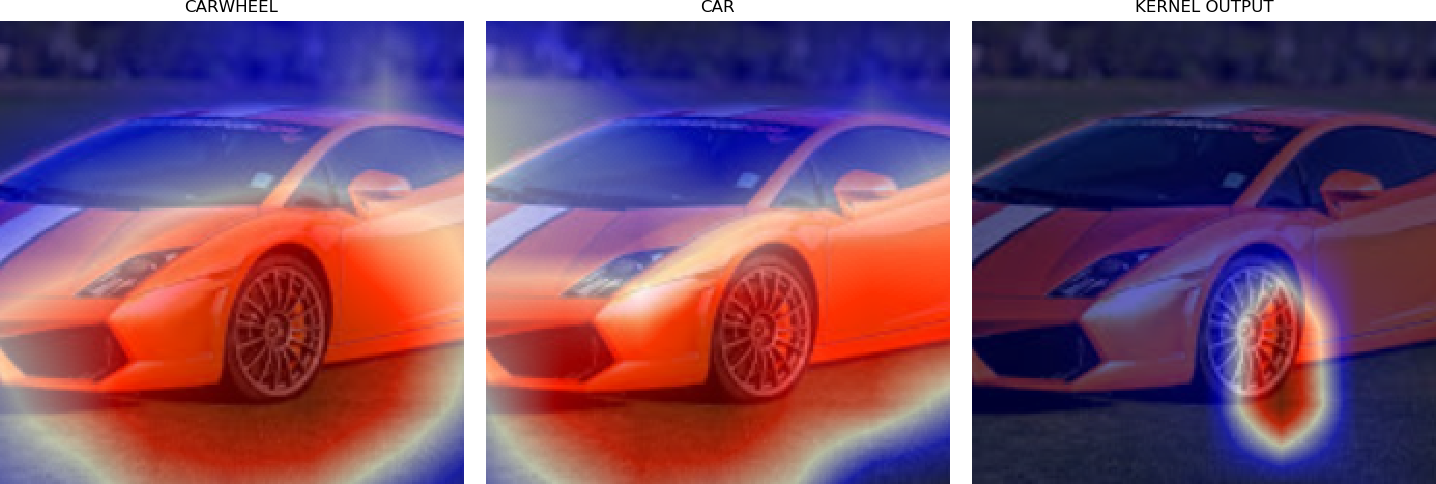}}%       
    }\\
     \caption{Results of our approach applied to a natural image dataset with overlapping classes.
       The class of interest is carwheel and car is the overlapping class.
       Each row contains triples of images with class activation maps superimposed on the original image.
       The first two images in each set show activation maps from the expert binary models,
       and the third image shows the output of our directed kernel function that
       localizes features relevant to the class of interest.
       The amplification parameter $\alpha$ was 5.}
    \label{fig:cars}
\end{figure*}

\subsection{Amplified Directed Divergence Kernel}
In order to extract overlapping features between two classes, we need a directed divergence or difference measure. For tensors $(x,x')$ we need a measure that will amplify only positive differences in activation values in $(x - x')$, because we are interested in recovering features in tensor $x$ with higher values than tensor $x'$, but not vice versa i.e. $K(x,x') \neq K(x',x)$. We introduce a kernel method called Amplified Directed Divergence Kernel (ADDK) that accepts two tensors $(x,x')$ of equal shape and returns another tensor of the same shape with amplified positive differences of $(x - x')$ as shown in \autoref{eq:addk}. The kernel method ensures that a maximum value of a given tensor is not zero in the normalization step. Normalization with maximum tensor values has shown promising empirical results, but we plan to explore other normalization techniques in the future.

\begin{equation}\label{eq:addk}
    K(x,x') = exp(\alpha(x/max(x) - x'/max(x')))
\end{equation}

The parameter $\alpha$ controls amplification of directed differences where
higher amplification will concentrate the resulting heat map to a smaller region.
To illustrate the kernel function operation, a simplified example with $\alpha=15$
is shown in \autoref{eqn:sampleOp}.

\begin{equation}
K\left(
\begin{bmatrix}
    1 & 1 & 5 \\
    0 & 6 & 4 \\
    0 & 1 & 0
\end{bmatrix},
\begin{bmatrix}
    8 & 0 & 7 \\
    1 & 4 & 3 \\
    1 & 2 & 1
\end{bmatrix}
\right)
= 
\begin{bmatrix}
    .0 & 12.2 & .5 \\
    .2 & 1808 & 79.8 \\
    .2 & .3 & .2
\end{bmatrix}
 \label{eqn:sampleOp}
\end{equation}

\noindent Tensors $x$ and $x'$ in $K(x,x')$ represent CAM outputs from respective binary expert models on the same image. Tensor sizes have been reduced and the result rounded for clarity.

\section{Experiments}

We apply our proposed dual-network technique to localize regions indicating COVID-19 in X-ray imagery.
However, due to the absence of localized and labeled bounding boxes for COVID-19  in X-rays,
the computed localizations cannot be easily validated.
Therefore, we also tested our technique on a natural imagery dataset for which the localizations
can be visually validated.

To train the expert binary models, we utilized transfer learning to mitigate the problem of training a robust image classifier with a small number of training samples from a novel class of interest.
We used a pretrained ResNet-152 architecture ~\cite{ResnetHe:7780459}
and replaced the final connected layer with appropriate classes and fine-tuned it with new data.
Stochastic gradient descent was used with a learning rate of 0.001 and momentum of 0.9.
Training was performed for 30 epochs and the best performing model based on validation accuracy was selected.

\begin{figure*}[t]
    \centering
    \subfloat[]{%
      \fbox{\includegraphics[width=0.90\columnwidth]{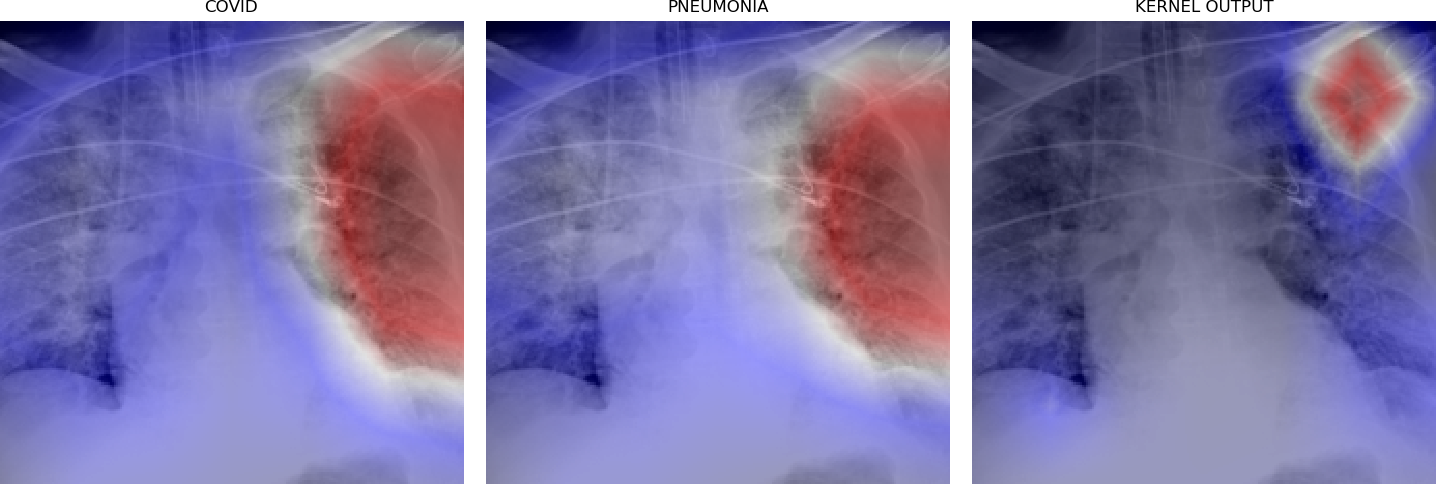}}
    }
    \hspace{0.1in}
     \subfloat[]{%
       \fbox{\includegraphics[width=0.90\columnwidth]{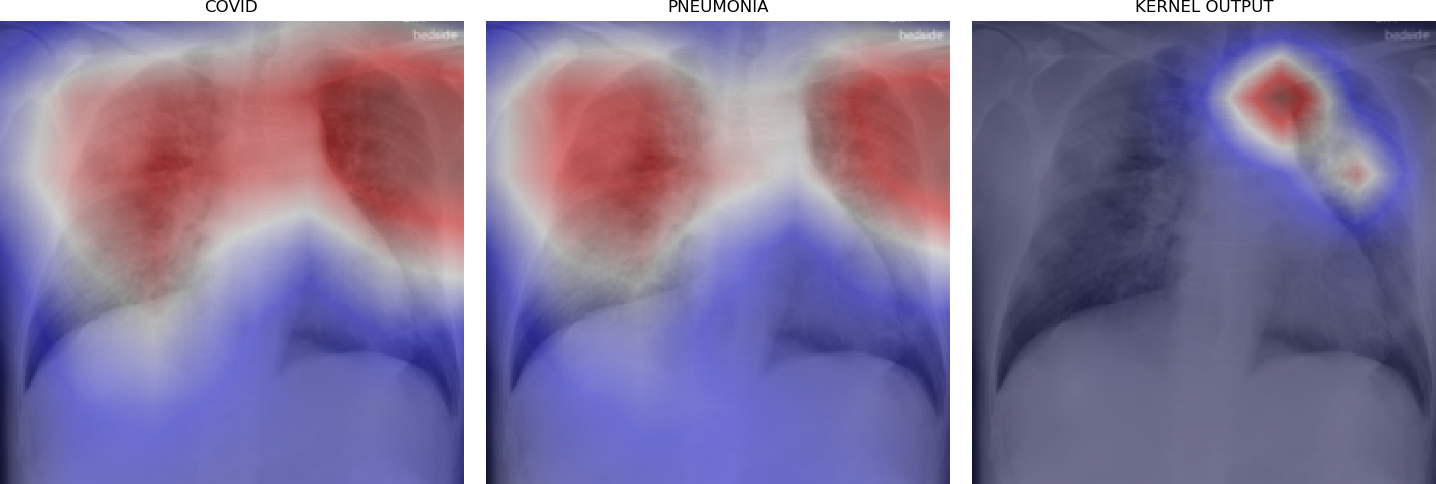}}%
    }\\
     \vspace{0.1in}
     \subfloat[]{%
       \fbox{\includegraphics[width=0.90\columnwidth]{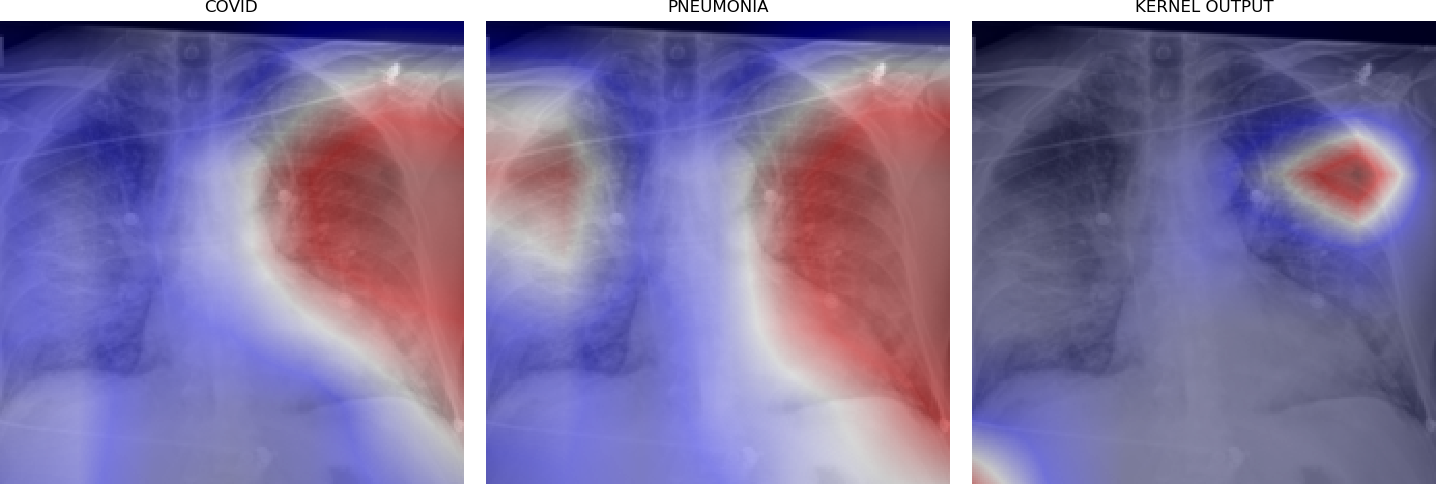}}
     }
     \hspace{0.1in}
     \subfloat[]{%
       \fbox{\includegraphics[width=0.90\columnwidth]{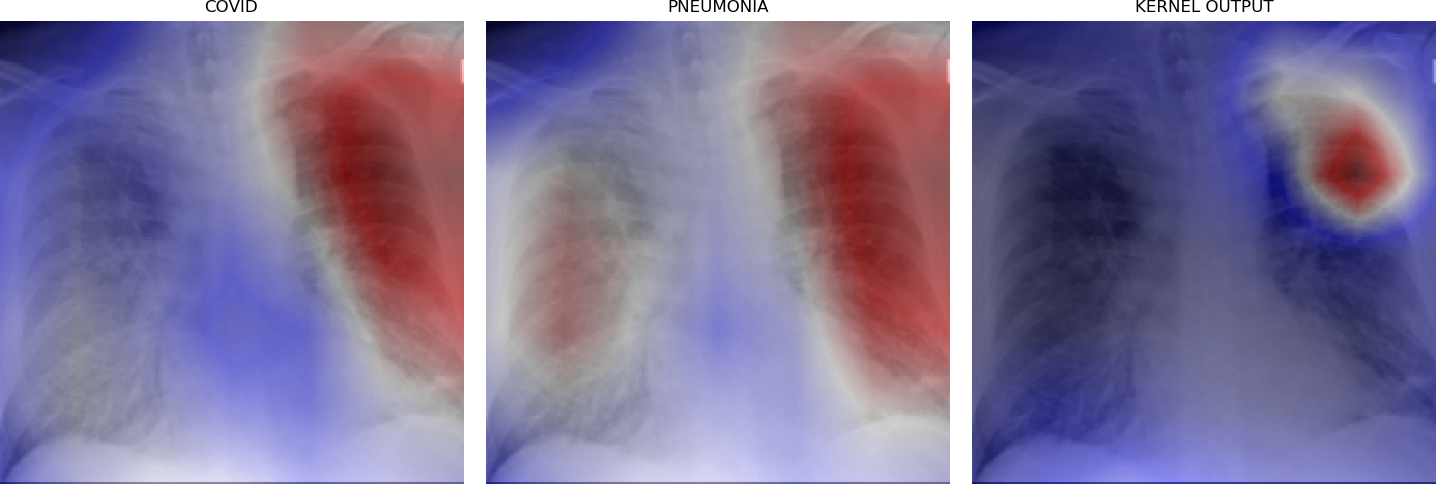}}%
    }\\
     \vspace{0.1in}
     \subfloat[]{%
       \fbox{\includegraphics[width=0.90\columnwidth]{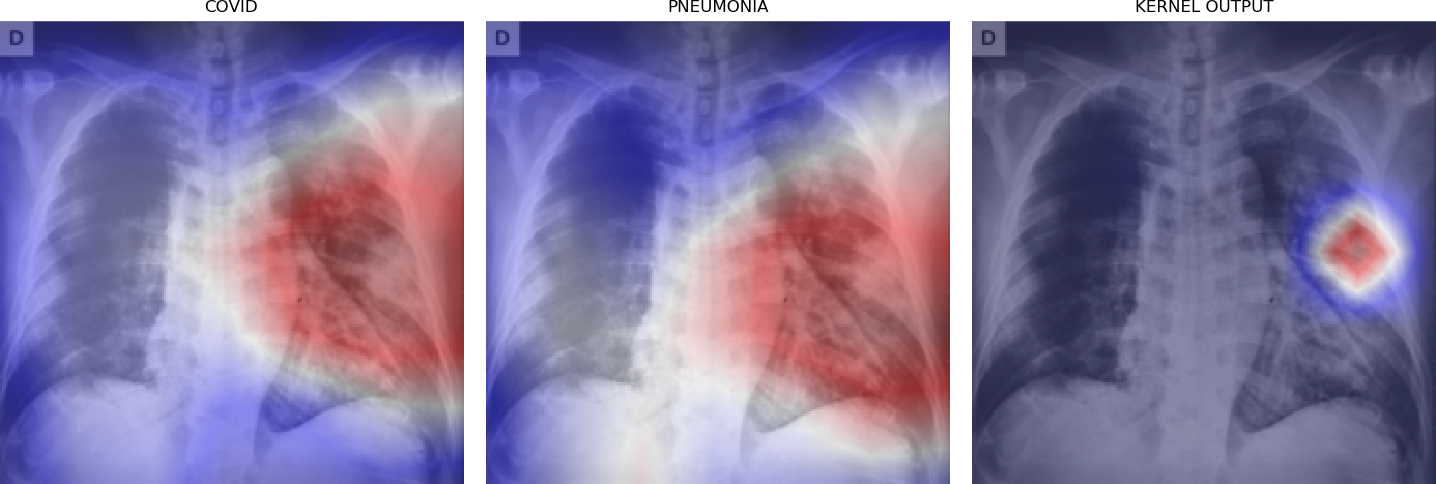}}%
     }
     \hspace{0.1in}
     \subfloat[]{%
      \fbox{\includegraphics[width=0.90\columnwidth]{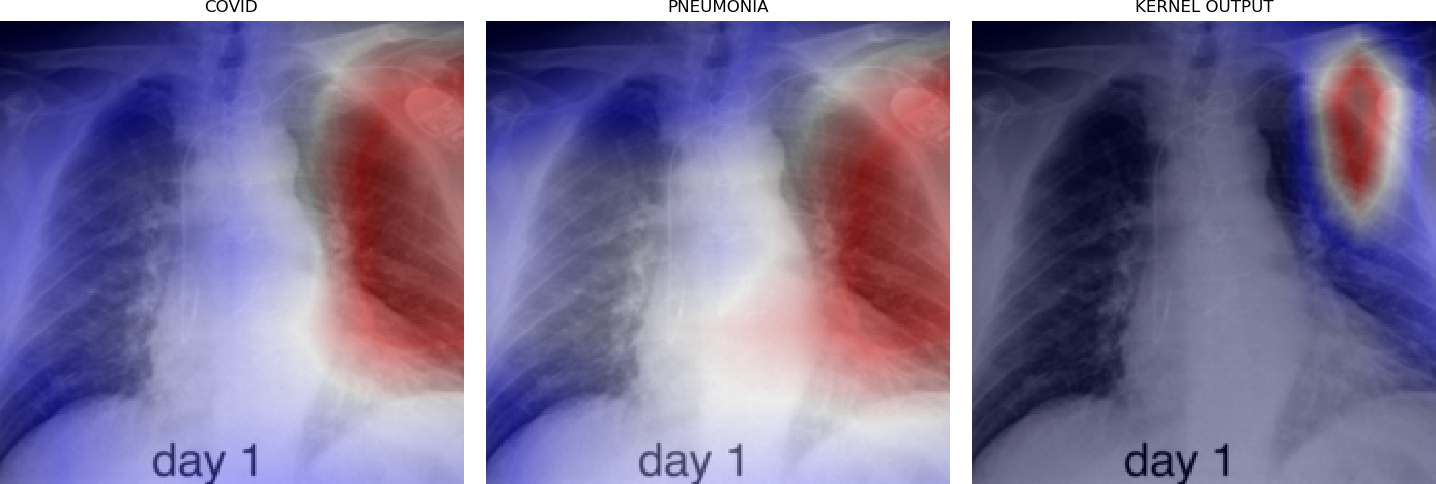}}%
    }         
     \caption{The results of our approach for distinguishing COVID-19 specific regions in X-ray imagery
       on publicly available X-ray datasets for COVID-19~\cite{cohen2020covid} and Pneumonia~\cite{Kermany2018}.
       In each set of three images, the first two show
       superimposed CAMs corresponding to the two classes predicted by the expert networks for COVID-19 and Pneumonia,
       respectively. The third image in each set is the output of our kernel function
       representing a heat map where regions in the image where the expert network for COVID-19
       was more confident have been localized.
     }
    \label{fig:results}
\end{figure*}

\subsection{Natural imagery}
We selected two categories of images with significant class overlap, viz. \emph{carwheel} and \emph{car},
the former being the  class of interest.
We fine-tuned two CNNs  pretrained on ImageNet as binary experts for car and carwheel.
The models were trained solely for classification and not for object localization.
Furthermore,  the carwheel expert model was fine-tuned with only thirteen images in order to
 simulate an environment with a novel class of interest that is likely to be data-starved (e.g. COVID-19).
The CAMs obtained from these two experts were passed through our novel  kernel function
to obtain a heat map that localizes regions in the image where the expert network for carwheel was more confident.

\autoref{fig:cars} shows results consisting of sets of three images; the first two are the CAMS
obtained from the carwheel and car expert networks respectively, and the 
third is the heat map computed using our kernel function that 
significantly improves the localization of the class of interest (carwheel).
This enhances the explainability of the classification decision in this class overlap scenario.

\subsection{Medical imagery\label{subsec:medimg}}
We utilized the COVID-19 chest X-ray dataset~\cite{cohen2020covid} and
 extracted COVID-19 samples with the posteroanterior view of the X-ray.
 The dataset by Kermany et. al~\cite{Kermany2018} was used for the Pneumonia X-ray images,
 which are also available from Kaggle.
Both data sets were processed into training, validation, and test splits using the  60/20/20 ratio.
\autoref{tbl:datasetsizes} shows the dataset sizes during training, validation, and testing.

\begin{table}[b]
  \caption{Sizes of the datasets used to train the binary expert models. Expression A/B represents the number of examples with
  positive and negative labels respectively.}
  \label{tbl:datasetsizes}
  \centering
  \normalsize
  %\large
  \begin{tabular}{|c|c|c|}
    \hline
        {~} & Train+Val  & Test\\ \hline
        COVID-19 & 79/140 & 27/47\\ \hline
        Pneumonia & 2563/949 & 856/317\\
    \hline
  \end{tabular}
\end{table}

\begin{figure}[]
    \centering
    \subfloat[$\alpha=15$]{%
      \label{subfig:alpha15}
      \fbox{\includegraphics[width=0.90\columnwidth]{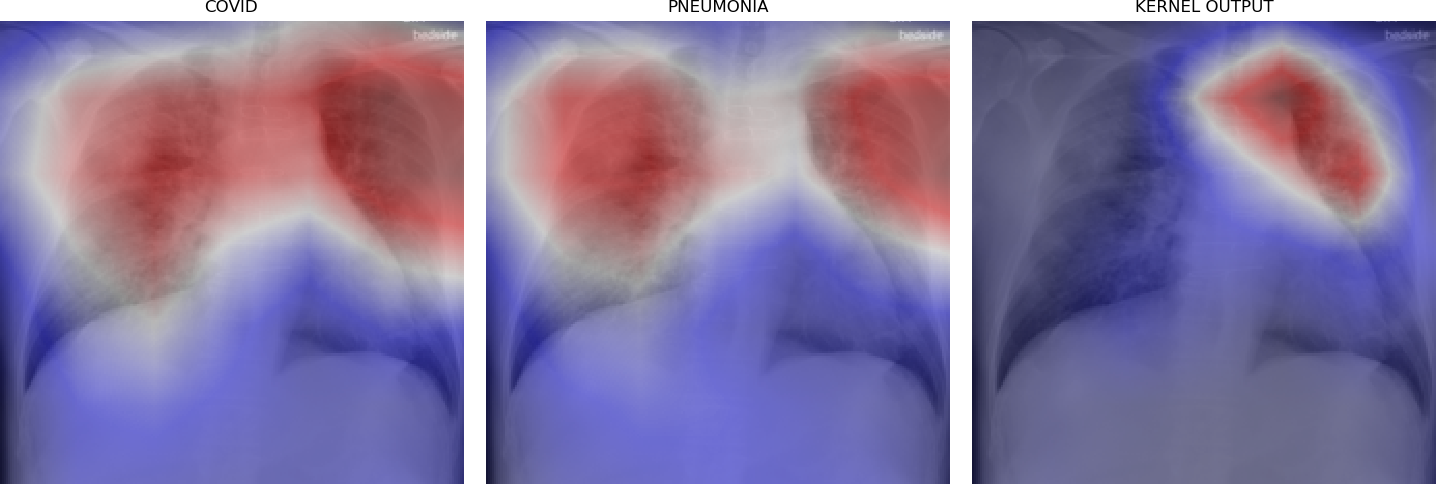}}%
    }
    % blank line to align
    \vspace{0.15in}
    \subfloat[$\alpha=50$]{%
      \label{subfig:alpha50}
       \fbox{\includegraphics[width=0.90\columnwidth]{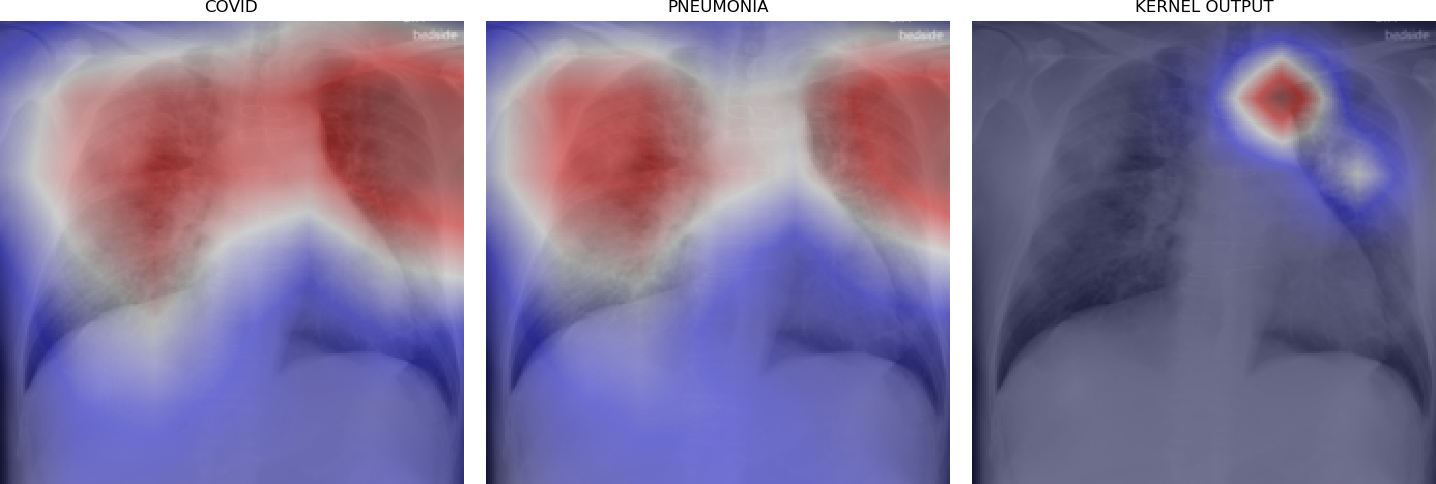}}%
    }
    \caption{Role of the amplification parameter in our kernel function (given by \autoref{eq:addk}).
      As the amplification parameter $\alpha$ is increased,
      the COVID-19 localization in the heat map output by the kernel function becomes more concentrated.
     }
    \label{fig:alphas}
\end{figure}

\autoref{fig:results} shows triples of X-ray images with superimposed class activation maps for predictions obtained from expert binary models (images one and two) with the third image showing the heat maps computed using our kernel.
The intended use of our method is to examine positive classifications from two possibly overlapping classes (i.e. COVID-19, Pneumonia) and extract discriminative features pertaining to the class of interest,
i.e. COVID-19. 
Triples \emph{(a)-(f)} show positive classifications of COVID-19 and Pneumonia by their respective binary expert models along with class activation maps that localize the image region responsible for that classification. The third image in each triple shows a better localized image region for COVID-19 as computed using our method.
Our method is intended to improve explainability of predictions under circumstances where both models return positive classifications resulting in significant overlap in activation maps.

\autoref{fig:alphas} demonstrates the role of the kernel parameter $\alpha$.
It controls amplification of the directed  differences among the activation maps.
Higher values of $\alpha$  concentrate the resulting heat map to a smaller region.

\section{Discussion}
We have described a novel method that improves predictive explainability of image classification by reducing uncertainty induced by class overlap.
For a classification task  with overlapping classes,
our approach creates multiple separate binary classification problems.
In this way, we avoid the uncertainty due to the overlap between classes.
Therefore, each binary model is allowed to become more confident for its specific task
which is reflected in its CAM.
A direct comparison of the CAMs allows us to localize image regions in the CAM
that explain why the image was classified in a  specific class of interest.
It should be noted that {\em our approach enhances explainability in settings with class overlap
  by enabling models trained solely for classification to be used for localization}.
This is {\em extremely useful in scenarios where the training
data was not annotated at the level of bounding boxes},
as is the situation for existing COVID-19 datasets.

Our  results show that the proposed method is effective in extracting and better localizing objects or regions associated with a class of interest that has significant overlap with another class. Furthermore, discriminative localization is performed without models having been explicitly trained for localization and object detection using labeled object bounding boxes. 

This work was motivated by our observations that numerous reported applications \cite{wang2020covid}
of image classification and object detection in support of rapid screening and diagnosis of COVID-19 were inhibited by the noisy data available to practitioners, which in turn increased uncertainty in associated predictions. We identified uncertainty induced by overlapping features of COVID-19 and non-SARS-CoV2 induced pneumonia.
We developed a method to mitigate uncertainty due to class overlap, that cannot be easily reduced
just by using more training data.
Our dual-network technique and our Amplified Directed Divergence Kernel function
helps domain experts, e.g. radiologists, in computer-aided diagnosis. 
For COVID-19 and regular pneumonia, which have shared symptoms,
our approach can help better isolate relevant regions in diagnostic imagery
that explain specific classifications.

\section{Conclusion and Future Work}
Our work improves the explainability of classification decisions
in scenarios with overlapping classes.
We do this by training more confident  models on simpler binary classification problems.
Our approach uses these simpler binary models for enhancing explainability by
improved localization, without training for localization.

We believe that our technique can be extended towards other useful applications.
An example is  monitoring training progress of classification models on data without localization ground truths, while having subject matter experts assess whether the model is discriminating proper or expected regions based on class. This is useful to asses that the model is learning some causal relationship between data and class rather than some spurious correlation induced by data artifacts, i.e. certain classes have some artificial mark produced by the collection process. Finally, by pairing subsequent evolutions of the same model, i.e. continuous retraining on new data, our method can extract shifts in activation maps induced by retraining and hence detect model drift or covariate shift of the data. 

We believe that our technique is promising in addressing uncertainty related to noisy data and further development will enable its use in numerous applications.
One direction of future work is to  investigate  image upsampling methods
in order to better map class activations to original imagery.
Whether variations of our kernel function can help improve explainability by better
localization can also be explored.
More experiments are needed to observe the effects of transformations used during training of expert models. For example, some domains such as natural images benefit from random geometric variations during training, while others, i.e. posteroanterior X-ray imagery, do not, as subject positioning is relatively constant between data points.
Finally, the effects of training expert models from scratch
instead of utilizing pretrained models that are fine-tuned can also be explored.

\bibliographystyle{IEEEtran}
\IEEEtriggeratref{11}
\bibliography{adk}

\end{document}